\documentclass[fleqn,10pt]{style/wlscirep}
\usepackage[utf8]{inputenc}
\usepackage[T1]{fontenc}

\usepackage{scalerel}
\usepackage{tikz}

\usepackage{mathtools, nccmath}

\DeclarePairedDelimiter{\nint}\lfloor\rceil

\usepackage{lineno}
\usepackage{hyperref}
\usepackage{graphicx}
\usepackage{multirow}
\usepackage{amsmath,amssymb,amsfonts}
\usepackage{amsthm}
\usepackage{mathrsfs}
\usepackage{xcolor}
\usepackage{textcomp}
\usepackage{manyfoot}
\usepackage{booktabs}
\usepackage{algorithm}
\usepackage{algorithmicx}
\usepackage{algpseudocode}
\usepackage{listings}
\usepackage{makecell}
\usepackage[retain-explicit-plus]{siunitx}
\usepackage{threeparttable}
\usepackage{xurl}

\usepackage{array}
\newcolumntype{P}[1]{>{\centering\arraybackslash}p{#1}}
\DeclareSIUnit\bar{bar}

\usepackage{blindtext}
\usepackage{hyperref}
\usepackage{nameref}

\newcounter{mylabelcounter}

\makeatletter
\newcommand{\labelText}[2]{%
\refstepcounter{mylabelcounter}%
\immediate\write\@auxout{%
 \string\newlabel{#2}{{\unexpanded{#1}}{\thepage}{{\unexpanded{#1}}}{mylabelcounter.\number\value{mylabelcounter}}{}}%
}
}
\makeatother

\newcommand\replyabs[2]{\labelText{#2}{#1}\textcolor{black}{#2}}

\makeatletter
\newcommand\footnoteref[1]{\protected@xdef\@thefnmark{\ref{#1}}\@footnotemark}
\makeatother

\title{An Explainable Machine Learning Framework for Railway Predictive Maintenance using Data Streams from the Metro Operator of Portugal\footnote{This version of the article has been accepted for publication, after peer review but is not the Version of Record and does not reflect post-acceptance improvements, or any corrections. The Version of Record is available online at: https://doi.org/10.1038/s41598-025-08084-1.}}

\author[1,*]{Silvia García-Méndez}
\author[1]{Francisco de Arriba-P\'erez}
\author[2]{Fátima Leal}
\author[3,4]{Bruno Veloso}
\author[4,5]{Benedita Malheiro}
\author[1]{Juan Carlos Burguillo-Rial}
\affil[1]{Information Technologies Group, atlanTTic, University of Vigo, Spain. }
\affil[2]{REMIT, Universidade Portucalense, Portugal}
\affil[3]{Faculty of Economics, University of Porto, Portugal}
\affil[4]{INESC TEC, Porto, Portugal}
\affil[5]{ISEP, Porto, Portugal}

\affil[*]{sgarcia@gti.uvigo.es}


\begin{abstract}
The public transportation sector generates large volumes of sensor data that, if analyzed adequately, can help anticipate failures and initiate maintenance actions, thereby enhancing quality and productivity. This work contributes to a real-time data-driven predictive maintenance solution for Intelligent Transportation Systems. The proposed method implements a processing pipeline comprised of sample pre-processing, incremental classification with Machine Learning models, and outcome explanation. This novel online processing pipeline has two main highlights: (\textit{i}) a dedicated sample pre-processing module, which builds statistical and frequency-related features on the fly, and (\textit{ii}) an explainability module. This work is the first to perform online fault prediction with natural language and visual explainability. The experiments were performed with the Metro\textsc{pt} data set from the metro operator of Porto, Portugal. \replyabs{the_results_are}{The results are above \SI{98}{\percent} for \textsc{f}-measure and \SI{99}{\percent} for accuracy. In the context of railway predictive maintenance, achieving these high values is crucial due to the practical and operational implications of accurate failure prediction. In the specific case of a high \textsc{f}-measure, this ensures that the system maintains an optimal balance between detecting the highest possible number of real faults and minimizing false alarms, which is crucial for maximizing service availability. Furthermore, the accuracy obtained enables reliability, directly impacting cost reduction and increased safety.} \replyabs{the_analysis_demonstrates}{The analysis demonstrates that the pipeline maintains high performance even in the presence of class imbalance and noise, and its explanations effectively reflect the decision-making process. These findings validate the methodological soundness of the approach and confirm its practical applicability for supporting proactive maintenance decisions in real-world railway operations.} Therefore, by identifying the early signs of failure, this pipeline enables decision-makers to understand the underlying problems and act accordingly swiftly.
\end{abstract}
\begin{document}

\flushbottom
\maketitle

\thispagestyle{empty}


\section{Introduction}

The digitalization of critical infrastructures and the adoption of Industry 4.0 technologies have driven profound transformations in sectors such as healthcare \cite{Sony2019}, manufacturing \cite{Oztemel2020,Jiang2021}, and transportation \cite{Li2023,Prakash2024}. In the latter, the massive deployment of sensors and monitoring systems has generated large volumes of data, allowing for continuous monitoring of asset status. This new scenario has led to the development of predictive maintenance (\textsc{p}{\footnotesize d}\textsc{m}) strategies, which aim to anticipate failures in real time by analyzing historical patterns. Compared to corrective approaches, \textsc{p}{\footnotesize d}\textsc{m} allows for intervention just before a breakdown occurs, thereby improving system availability, reducing operating costs, and enhancing safety standards.

Online monitoring generates large volumes of heterogeneous sensor data at a high pace, falling into the category of big data \cite{Kolajo:2019,Vial2022}. These endless data streams, when adequately mined, can anticipate failures and support decision-making in multiple industry domains \cite{Wang2020,ZONTA2020}, playing an essential role in the implementation of \textsc{p}{\footnotesize d}\textsc{m} \cite{Sahal2020}. By anticipating failures, \textsc{p}{\footnotesize d}\textsc{m} allows companies to increase productivity and reduce operational costs \cite{Dalzochio2020}. In the railway context, characterized by dynamic environments, high-reliability demands, and components subject to constant wear, the effective implementation of predictive solutions represents a key opportunity to optimize the operation and maintenance of critical transportation infrastructures.

In the public passenger transportation domain (air, rail, and road), failures cause delays and cancellations, lower the reputation, and increase the costs of the involved operators \cite{Irannezhad2020,marzouk2024framework}. Specifically, in railway transports, wheel and door failures are a significant cause of delays \cite{Sahal2020}. Nowadays, trains generate countless sensor data streams, \textit{e.g.}, light, position, pressure, temperature, or vibration, which online predictive algorithms can use. Data stream analysis continuously monitors and detects relevant changes in the status of components and systems, alerting decision-makers to take action. Therefore, data stream processing is essential for real-time \textsc{p}{\footnotesize d}\textsc{m}, namely, for Intelligent Transportation System (\textsc{its}) applications to make transportation networks safer and more sustainable \cite{NEILSON2019,Zargayouna2020}.

In this context, \textsc{p}{\footnotesize d}\textsc{m} relies extensively on Machine Learning (\textsc{ml}) techniques. However, many of the employed techniques are opaque \cite{Burkart2021,Vollert2021}, leaving decision-makers unclear about the reasoning behind them. Transparent predictive techniques, whether interpretable or explainable, provide intelligible learning results. In this context, interpretable models are those whose reasoning can be easily understood by humans, such as decision trees, regression, or rule-based predictors. Conversely, black box models, like support vector machines or deep neural networks, require ad hoc explainability methods \cite{Adadi2018,Spadon2019}. Therefore, early detection and explanation of faults allow maintenance experts to identify the underlying problems and apply corrective measures.

In the specific case of the railway sector, the focus of our research, \textsc{p}{\footnotesize d}\textsc{m} has assumed a crucial role to enhance service reliability, reduce operating costs, and improve system safety \cite{Binder2023}. In this sense, existing solutions aim to continuously monitor data from onboard sensors to anticipate failures, enabling efficient planning of maintenance tasks and preventing unexpected service downtime. Specifically, the works in state-of-the-art have been developed to detect anomalies in critical components, including air production units, wheels, doors, brakes, and electrical systems, among others \cite{Davari2021,H-Nia2024}. However, the most recent work in this area has incorporated \textsc{ml} techniques applied to real-time data to identify behavioral patterns that precede operational failures \cite{Meira2023}. In contrast, there also exist solutions that approach the problem from a non-classification point of view \cite{Davari2021,Le-Nguyen2021,Le-Nguyen2023realtime,Le-Nguyen2023}. However, despite these advances, the field continues to face significant challenges, including the lack of interpretability in models, the limited integration of real-time explanations, and the need to adapt methods to variable conditions and scenarios with highly unbalanced classes. This has motivated the development of more transparent approaches that can combine predictive accuracy with operational explainability, as in our case, thereby facilitating their practical adoption in real-world railway environments.

By classifying sensor data streams on the fly with the help of \textsc{ml} algorithms and explainability techniques, the current work aims to address both factors simultaneously. Overall, it enhances transportation operators' quality of service, reputation, productivity, and profits through real-time natural language and visual explainability. The ultimate objective of this research is to mine the incoming train sensor data stream from a public railway operator to anticipate failures and support maintenance decisions. The solution exploits a transparent, data-driven \textsc{p}{\footnotesize d}\textsc{m} method for the public transportation sector. It is composed of three main modules: (\textit{i}) data pre-processing, encompassing feature engineering, analysis, and selection; (\textit{ii}) stream-based failure classification; and (\textit{iii}) failure explanation supported by \textsc{ml} algorithms \cite{Molnar2020,Pang2023} and Natural Language Processing (\textsc{nlp}) techniques \cite{Søgaard2021}. Therefore, compared to the literature, the proposed system contributes a novel real-time data-driven \textsc{p}{\footnotesize d}\textsc{m} solution, comprising dedicated sample pre-processing and classification, together with textual and visual outcome explanations for both \textsc{ml} and maintenance experts. The method was tested using the Metro\textsc{pt} data set collected from trains in Metro do Porto, Portugal. The empirical results with the Adaptive Random Forest Classifier exhibit competitive accuracy and micro and macro $F$-measure values. Moreover, each incoming event is classified, and its class label is explained by detailing the sensors' abnormal behavior and other relevant features.

The rest of this paper is organized as follows. Section~\ref{sec:2} reviews the \textsc{p}{\footnotesize d}\textsc{m} background, focusing on stream-based and explainable \textsc{ml}. Section~\ref{sec:proposed_method} describes the proposed method composed of data processing, classification, and explainability modules. Section~\ref{sec:experimental_results} presents the experiments performed and the results of the empirical evaluation. Finally, Section~\ref{sec:conclusion} summarizes and discusses the outcomes of this work.

\section{Related work}
\label{sec:2}

The Internet of Things and Industry 4.0 has contributed to the exponential increase in data generated by devices, machines, manufacturing lines, and industrial processes \cite{Lampropoulos2019,Katsaros2023}. This big data scenario presents new challenges to researchers. The current work, which focuses on the processing of such data streams for \textsc{p}{\footnotesize d}\textsc{m}, aims to identify and explain early signs of failure. 
Early detection of a failure improves process management and reduces the associated economic, environmental, and social costs \cite{Carvalho2019,Proto2020}. In the railway sector, \textsc{p}{\footnotesize d}\textsc{m} reduces operational costs and train downtime as well as boosts the quality of service.

The literature identifies four maintenance categories \cite{ZONTA2020}: (\textit{i}) corrective -- maintenance takes place after the fault has occurred; (\textit{ii}) preventive -- uses time slots to replace components; (\textit{iii}) predictive -- performs the early detection of failures; and (\textit{iv}) prescriptive -- provides valuable suggestions for extending the life of the equipment. As a data-intensive approach, \textsc{p}{\footnotesize d}\textsc{m} requires specialized pre-processing techniques to resolve inconsistencies and redundancies and prepare the highly heterogeneous data for the subsequent phases \cite{Bekar2020}. Moreover, according to Xie, J. \textit{et al.} (2020) \cite{xie2020systematic}, as most approaches are based on opaque models, maintenance professionals remain unaware of how the results were obtained.

\subsection{Predictive maintenance for the railway sector}

Existing surveys on \textsc{p}{\footnotesize d}\textsc{m} for railway operators analyze and compare multiple data-driven approaches \cite{davari2021survey,Gama:2022}, including the prediction of failures as well as of the remaining service life. Most of these works employ classical \textsc{ml} algorithms, such as decision trees, neural networks, outlier detectors, regressors, and rule-based methods (see Table \ref{tab:vdetection}). The latter suggests designing online processing pipelines that comprise dedicated pre-processing and transparent predictive algorithms to help maintenance specialists make well-informed decisions. This related work compares the most recent \textsc{p}{\footnotesize d}\textsc{m} works in terms of: (\textit{i}) data pre-processing; (\textit{ii}) stream-based classification; and (\textit{iii}) transparency.

\begin{description}
    \item \textbf{Pre-processing} comprises tasks such as data cleansing, feature engineering, information fusion, data balancing, feature selection, and feature analysis. Pre-processing is a time-consuming task for knowledge discovery in the context of a data stream. Ramírez-Gallego, S. \textit{et al.} (2017)\cite{ramirez2017survey} analyze pre-processing techniques for stream-based applications, contemplating concept drifts, data reduction, ensemble learning, and sliding windows. \textsc{p}{\footnotesize d}\textsc{m} requires dedicated online feature engineering methods for failure detection. In this context, the literature encompasses stream-based data manipulation approaches that summarize behavior frames using tuples of statistical data \cite{MANCO2017} or segment the data into intervals to generate features from the analog and digital sensors \cite{Davari2021}. 

    \item \textbf{Data-driven predictive maintenance} monitors incoming sensor data in real time to establish mechanical or equipment indicators by exploiting advanced \textsc{ml} methods to dynamically detect functioning patterns and operating conditions \cite{Gama:2022}. \textsc{p}{\footnotesize d}\textsc{m} has been explored in the streaming scenario for:
    \begin{itemize}
        \item \textbf{Anomaly detection} using outlier detection methodologies \cite{MANCO2017}. 
        \item \textbf{Estimation of the remaining useful life} of rail axle bearings through Online Support Vector Regression (\textsc{osvr}) \cite{FUMEO2015}. To address the big data scenario, the authors propose a heuristic trade-off between accuracy, runtime, and resources required by \textsc{ml} models.
        \item \textbf{Failure prediction} trough sparse autoencoders, using unsupervised methods based on Deep Learning \cite{Davari2021,Ribeiro:2023}.
    \end{itemize}
 
    \item \textbf{Transparent} \textsc{p}{\footnotesize d}\textsc{m} models help decision-makers to understand the need to replace or repair a particular component. Predictive \textsc{ml} models can be intrinsically interpretable (\textit{e.g.}, decision trees, graph-based approaches, regressors, and rule-based methods) or opaque (\textit{e.g.}, neural networks) \cite{Pazera2019}. Interpretable models allow users to choose whether or not to trust them, reduce bias, and help discover additional insights \cite{Longo2020}. 
    Graph-based approaches can achieve promising performance on \textsc{p}d\textsc{m} perception tasks by revealing the dependency relationship among parts and components of the equipment. 
    Opaque models can be coupled with post hoc techniques for explainability purposes. 
    Explainability techniques can be model-agnostic or model-specific and can be either global (explaining the entire model) or local (explaining each prediction). Examples of local post hoc model-agnostic techniques are the Local Interpretable Model-agnostic Explanations (\textsc{lime}) \cite{Ribeiro:2016} and SHapley Additive exPlanations (\textsc{shap}) \cite{Lundberg:2017}. Alternatively, explanations can be classified under the example-based, feature-importance, knowledge-extraction, and visual categories \cite{Vollert2021}. The main explainable \textsc{p}{\footnotesize d}\textsc{m} works found in the literature are listed below.
    
    \begin{itemize}
        \item Manco, G. \textit{et al.} (2017) \cite{MANCO2017} propose an unsupervised fault explanation system based on features with abnormal values.
    
        \item Allah Bukhsh, Z. \textit{et al.} (2019) \cite{ALLAHBUKHSH2021} present an explainable classifier for railway switches using decision trees, random forests, and gradient-boosted trees. They adopt the \textsc{lime} technique to identify the features that contribute positively or negatively to each class label. 
 
        \item Ribeiro, R. P. \textit{et al.} (2023) \cite{Ribeiro:2023} describe a neural symbolic explainer, which combines oversampling with the \textsc{amr}ules algorithm to describe anomalies detected by an autoencoder.
    \end{itemize}
 
\end{description}

As previously mentioned, recent \textsc{p}{\footnotesize d}\textsc{m} solutions seek continuous monitoring of data, regardless of the sector. A representative example is the work by Niu, X. \textit{et al.} (2021)\cite{niu2021data}, which focused on the airline flight delays problem and addressed the challenge using data-driven dynamics, similar to our study. Within the railway sector, authors have proposed stream-based monitoring, anomaly detection, and anomaly prediction, as seen in the work by Le-Nguyen, M. H. \textit{et al.} (2021)\cite{Le-Nguyen2021}, which led to the development of the InterCE system and subsequent investigations \cite{Le-Nguyen2023,Le-Nguyen2023realtime} for the passenger access system, among others. However, the latter works addressed \textsc{p}{\footnotesize d}\textsc{m} from a non-classification perspective. The sole exception is the study by Meira, J. \textit{et al.} (2023)\cite{Meira2023}. The authors presented an unsupervised real-time anomaly detection system.

\subsection{Research contribution}

The research presented in this paper is based on a combination of theoretical and technical foundations that converge on the need for effective, explainable \textsc{p}{\footnotesize d}\textsc{m} solutions adapted to real-time data processing. First, it draws on existing empirical evidence demonstrating the positive impact of \textsc{p}{\footnotesize d}\textsc{m} on reducing costs, downtime, and critical failures in railway systems. Second, it builds on the accumulated knowledge of \textsc{ml} techniques applied to data flow analysis, which enables the construction of incremental models capable of adapting to dynamic environments. Furthermore, this research draws on the growing interest in the explainability of models, integrating mechanisms that not only predict failures with high accuracy but also justify their decisions through understandable textual and visual descriptions. Note that the work is based on real data obtained from the railway operator in Portugal, which ensures that the proposed solution responds to real operating conditions.

In short, we propose a novel, transparent, and operational solution for real-time \textsc{p}{\footnotesize d}\textsc{m} in the railway sector. The proposed method encompasses stream-based data pre-processing, failure classification, and failure explanation supported by \textsc{nlp} techniques. Its main contributions are articulated around three interdependent axes: (\textit{i}) processing, (\textit{ii}) prediction, and (\textit{iii}) explainability. First, a pre-processing module is introduced that is capable of dynamically generating statistical and frequency characteristics by applying \textsc{fir} filters to sliding windows adapted to the nature of the analog and digital signals in the input data. Second, an incremental classification architecture is applied based on explainable models optimized under a prequential protocol and variance thresholding dimensionality reduction. Third, the explainability module works with global feature-relevance descriptions (textual) and local feature-relevance descriptions (textual and visual), thereby generating automatic natural language and interpretable visualizations that explain the causes of each prediction. The solution was experimentally validated on the MetroPT dataset by Metro do Porto, the operator. The promising results obtained not only significantly exceed the values reported in previous works with the same dataset but also confirm the robustness and practical applicability of the proposed methodology.

While the literature presents a significant number of \textsc{p}{\footnotesize d}\textsc{m} approaches for the industry in general, scant research has been produced on the online pre-processing, classification, and explanation of railway data. Other approaches have proven effective in controlled or offline environments, achieving good levels of accuracy in identifying anomalies. Among their main advantages is the ability to model complex, nonlinear relationships in the data, which is useful for detecting subtle patterns of degradation. However, many of these proposals have significant limitations in real-life scenarios. In particular, they often lack real-time processing capabilities, making them difficult to apply directly in continuous operating environments, such as railways. It is the case of the works by Manco, G. \textit{et al.} (2017) \cite{MANCO2017}, Allah Bukhsh, Z. \textit{et al.} (2019) \cite{ALLAHBUKHSH2021}, and Davari, N. \textit{et al.} (2021) \cite{Davari2021}. Furthermore, many of these models operate as black boxes, offering no comprehensible explanations for the decisions made, which reduces their acceptability by maintenance technicians, as in Fumeo, E. \textit{et al.} (2015) \cite{FUMEO2015} and Meira, J. \textit{et al.} (2023) \cite{Meira2023}.

Table \ref{tab:vdetection} compares the research that addresses online and/or transparent \textsc{p}{\footnotesize d}\textsc{m} for the railway sector. In light of the latter comparison, this work is the first to perform fault prediction in streaming mode, combining natural language and visual explainability.

\begin{table*}[tbh]
 \centering
 \small
 \caption{Comparison of \textsc{p}{\footnotesize d}\textsc{m} research.}~\label{tab:vdetection}
 \begin{tabular}{ccc@{}c@{}cc}
    \toprule
    \textbf{Authorship} & \textbf{Objective} & \textbf{Data processing} & \textbf{Decision technique} & \textbf{Decision processing} & \textbf{Explainability} \\
    \midrule
    \cite{MANCO2017} & Fault prediction & Offline & Outlier Detector & Offline & Feature importance \\
    \cite{Davari2021} & Fault detection & Offline & Sparse Variational Autoencoder & Offline & -- \\
    \cite{ALLAHBUKHSH2021} & Maintenance prediction & Offline & Tree-based classifiers & Offline & Feature importance \\
    \cite{FUMEO2015} & Fault detection & \textsc{na} & Online Support Vector Regressor & Online & -- \\
    \cite{Ribeiro:2023} & Fault detection & Online & Sparse Autoencoder & Offline & Natural language \\
    \cite{Meira2023} & Maintenance prediction & Online & XStream & Online & --\\
    \midrule 
    \multirow{1}{*}{\textbf{Current}}  & \multirow{2}{*}{Fault prediction} & \multirow{2}{*}{Online} & \multirow{2}{*}{Explainable classifiers} & \multirow{2}{*}{Online} & Natural language \\
    \multirow{1}{*}{\textbf{Proposal}} & & & & & \& visual\\
    \bottomrule
 \end{tabular}
\end{table*}

\section{Proposed method}
\label{sec:proposed_method}

\begin{figure*}[tbh]
\centering
\includegraphics[scale=0.12]{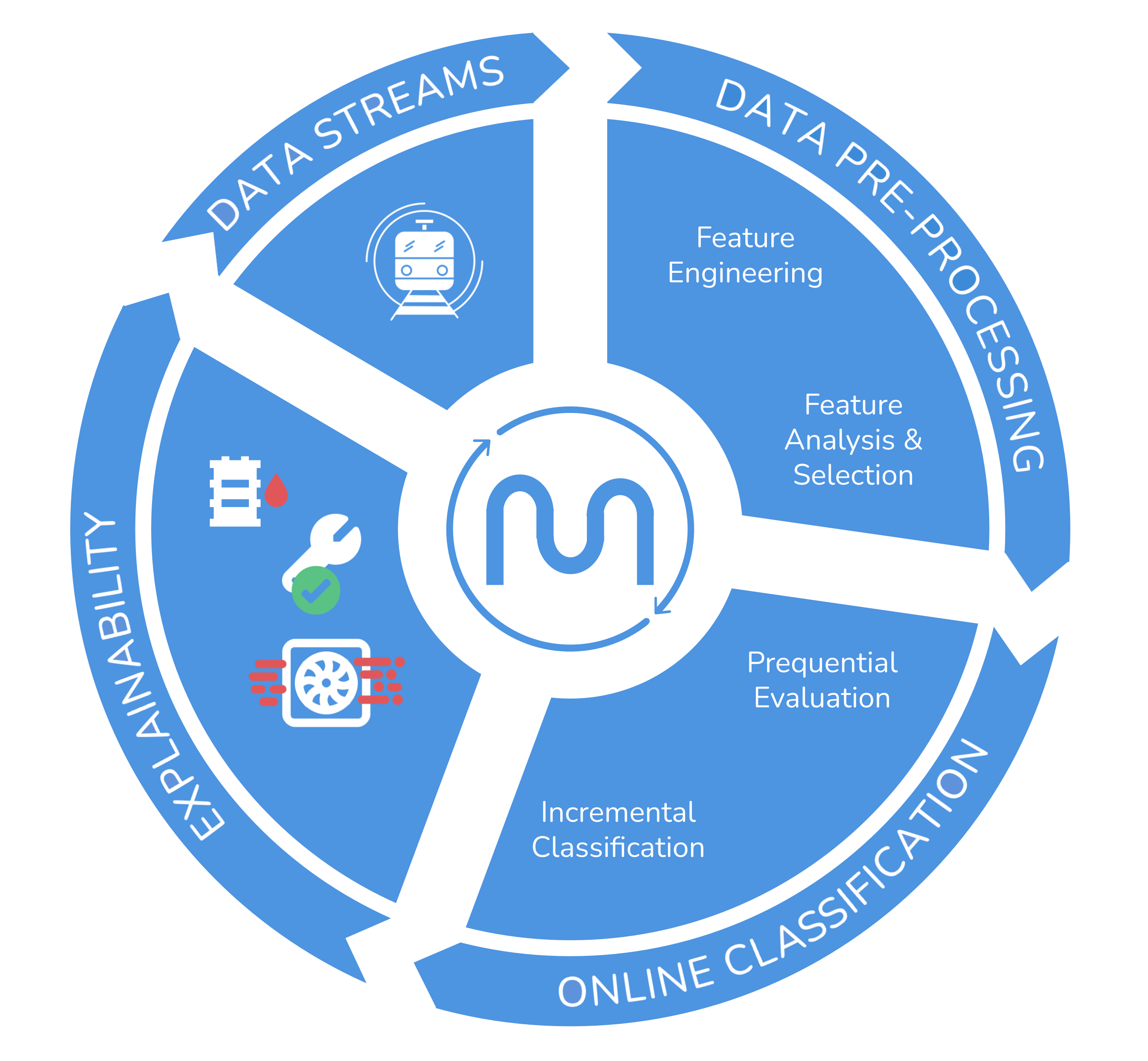}
\caption{\label{fig:scheme}Transparent online \textsc{p}{\footnotesize d}\textsc{m} cycle.}
\end{figure*}

The online \textsc{p}{\footnotesize d}\textsc{m} cycle is presented in Figure \ref{fig:scheme}. It is composed of three main modules: (\textit{i}) data pre-processing (Section \ref{sec:data_processing}), (\textit{ii}) \textsc{ml} classification (Section \ref{sec:classification}), and (\textit{iii}) explainability (Section \ref{sec:explainability}). All three modules work over data streams.

\subsection{Online data pre-processing}
\label{sec:data_processing}

Data pre-processing is crucial to ensure high-quality input data and, consequently, accurate classification results. This module comprises feature engineering, feature analysis, and selection procedures. 

\subsubsection{Feature engineering}
\label{sec:feature_engineering}

Sensor data, especially analog signals, may exhibit random oscillations from sample to sample, corresponding to white noise. 
To remove this undesirable noise, the analog signals are processed by four sliding-window finite impulse response (\textsc{fir}) filters \cite{Plonka2018,Segovia2018}. A sliding-window \textsc{fir} filter transforms an original analog signal into a white-noise-free signal as long as its length is adequate \cite{Arumugam2020}. Thus, the initial step is to automatically determine the minimal number of samples or sliding window length of each \textsc{fir} filter using a subset of the first two days of data:

\begin{itemize}

    \item Determine the relative minimum values of each feature in the experimental data set.
    \item Compute the number of samples between relative minimum values of each feature.
    \item Merge the number of samples of each feature into a list.
    \item Compute, using the previous list, the average and quartile distribution of the following four intervals: (\textit{i}) from minimum to the first quartile (\textsc{q\textsubscript{1}}), (\textit{ii}) from \textsc{q\textsubscript{1}} to the median, (\textit{iii}) from median to the third quartile (\textsc{q\textsubscript{3}}), and (\textit{iv}) from \textsc{q\textsubscript{3}} to the maximum.
    \item Define the size of the sliding windows of the \textsc{fir} filters based on the average, \textsc{q\textsubscript{1}}, \textsc{q\textsubscript{2}}, \textsc{q\textsubscript{3}} values. 
 
\end{itemize}

\begin{figure*}[htbp]
\centering
\includegraphics[scale=0.13]{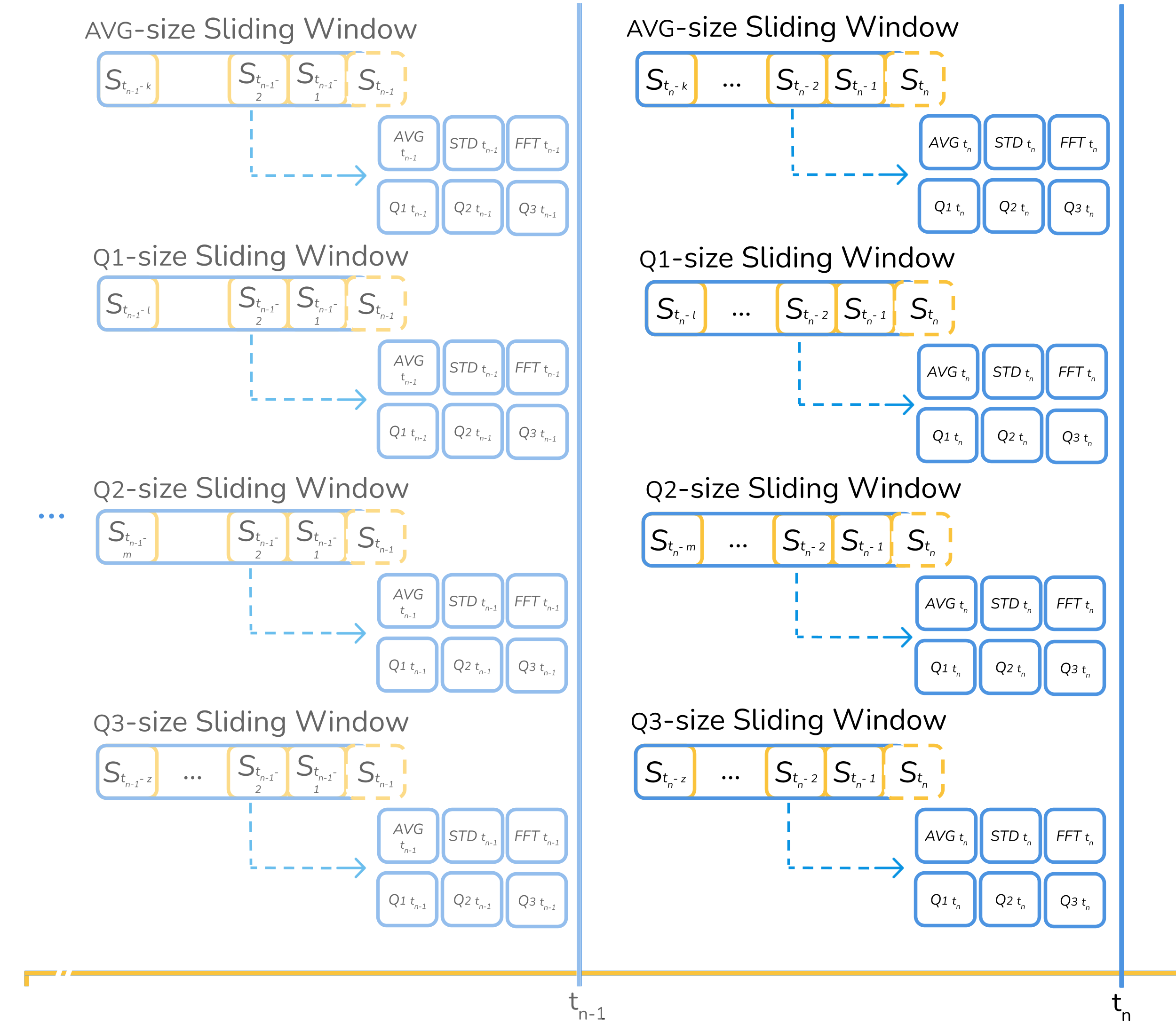}
\caption{\label{fig:windows}Features engineered using the \textsc{fft} and \textsc{fir} filters.}
\end{figure*}

The resulting sliding window lengths correspond to the number of samples required to overcome a cold start for each filter. Then, for each sample in the experimental data set, six new features are engineered per window filter -- average, standard deviation, \textsc{q\textsubscript{1}}, \textsc{q\textsubscript{2}}, \textsc{q\textsubscript{3}} and Fast Fourier Transform (\textsc{fft}\footnote{The \textsc{fft} represents the signal in the frequency domain \cite{Plonka2018}.}) -- according to Equation \ref{eq:feature_engineering}, where $n$ represents the identifier of the last incoming sample and $X[n]$ the vector holding the values of a particular feature. Note that these statistics were selected based on experimental tests. The first five engineered features characterize the signal distribution function, whereas the last detects anomalous frequencies, all within the sliding window of each \textsc{fir} filter. This process is illustrated in Figure \ref{fig:windows}.

\begin{equation}\label{eq:feature_engineering}
    \begin{aligned}
    \color{black}
    X[n] = \{ x[n-w+1],\ldots,x[n]\}. \\
    \color{black}
    Y[n] = \{y_0[n], y_1[n],\ldots,y_{w-1}[n]\} \mid \\ 
    \color{black}
    y_0[n]\leq y_1[n]\leq\ldots\leq y_{w-1}[n], \\
    \color{black}
    \mbox{where} \; \forall x \in X[n], \; x \in Y[n]. \\ \\
    \color{black}
    avg^w[n]=\frac{1}{w}\sum_{i=0}^{w} y_i [n]\\
    \color{black}
    std^w[n]=\sigma(X[n]) \\
    \color{black}
    Q^w_{1}[n]=y_{\nint{\frac{1}{4}w}}[n] \\
    \color{black}
    Q^w_{2}[n]=y_{\nint{\frac{2}{4}w}} [n]\\
    \color{black}
    Q^w_{3}[n]=y_{\nint{\frac{3}{4}w}} [n]\\
    \color{black}
    F^w[n]=|FFT(X[n])|
    \end{aligned}
\end{equation}

Consequently, the proposed solution presents a robust configuration against errors or outliers. Each sliding window reduces signal noise at different frequencies and interpolates the absence of data. Moreover, quartiles in the sliding windows and the engineered features purge spurious sensor measurements.

\subsubsection{Feature analysis and selection}
\label{sec:feature_analysis_selection}

In the case of stream-based \textsc{ml} classification, reducing the number of features by removing those less relevant to the task is essential. The chosen feature analysis and selection technique -- variance thresholding -- relies on feature variance. It assumes that features with low variance are less likely to be relevant than features with high variance \cite{Cao2020,Treistman2022}. This filtering technique first calculates the variance of each feature and then selects the features whose variance meets an experimentally defined threshold.

\subsection{Online classification}
\label{sec:classification}

The stream-based \textsc{ml} models were selected based on their interpretability and promising performance in similar classification problems \cite{Davari2021,Ribeiro:2023}.

\begin{itemize}
 
    \item \textbf{Gaussian Naive Bayes} (\textsc{gnb}) \cite{Xue2021} is a variant of the Naive Bayes (\textsc{nb}) model that processes streams by following a Gaussian normal distribution.
 
    \item \textbf{Hoeffding Tree Classifier} (\textsc{htc}) \cite{Pham2017} is a single tree induction model that exploits small amounts of samples to perform optimal splitting.
 
    \item \textbf{Hoeffding Adaptive Tree Classifier} (\textsc{hatc}) \cite{Stirling2018} is a single tree model that monitors branch performance to replace branches with decreasing performance by new and more accurate branches.
 
    \item \textbf{Adaptive Random Forest Classifier} (\textsc{arfc}) \cite{Gomes2017} is an ensemble of tree models that induces diversity, using re-sampling and randomly selecting feature subsets for node splits and applies concept drift detectors per base tree. The classification outcome results from voting, where votes are weighted based on the test-then-train accuracy of the forest trees.
 
\end{itemize}

Based on sliding windows, the adopted feature engineering technique enables these models to create non-stationary knowledge spaces, where outdated information is forgotten, compared to existing methods (see Table \ref{tab:vdetection}). To optimize the results and save time, the original data set is downsampled by a factor of 500, and the resulting random subset is used to set the hyperparameters of the classifiers.

Algorithmic performance was evaluated using several standard classification metrics. First, accuracy was calculated, which reflects the proportion of correctly classified instances relative to the total number of samples analyzed. In addition, \textsc{f}-measure was employed in the macro-averaging and micro-averaging versions. While macro-averaging \textsc{f}-measure assigns equal weight to each class, allowing for the evaluation of model performance in situations of class imbalance, micro-averaging \textsc{f}-measure weights classes according to their frequency, thus providing a comprehensive view of performance based on the actual volume of each class. Finally, runtime data were included to quantify the computational efficiency of the models, a fundamental aspect in classification systems with continuous data flow. All these metrics were calculated using the prequential evaluation protocol \cite{Gama2013}, which involves evaluating the model sequentially as it receives new samples. This protocol first applies the prediction to the current sample and subsequently uses that sample to update the model, thereby emulating the operation of an online learning system.

\subsection{Online explainability}
\label{sec:explainability}

Natural language descriptions of the advent of a fault and its cause are highly relevant for the maintenance team. Thus, the information displayed has a dual purpose:

\begin{itemize}
    \item Present the sensors that exhibit abnormal behavior textually and visually.
 
    \item Detail the most relevant features involved in the classification, including the most relevant \textsc{fir} filters, using intelligible descriptions.

\end{itemize}

To extract this information, the decision path of the estimator (one estimator in the case of \textsc{htc} and \textsc{hatc} and a configurable number of estimators in the case of the \textsc{arfc}, see hyperparameter \texttt{\small model} in Listing \ref{arfc_conf}) is traversed to gather the involved features that meet the \textit{greater than} condition together with their frequency (number of occurrences). Then, they are ordered by decreasing frequency to extract the five most relevant. Furthermore, the \textsc{ml} model is also described through the selected relevant features: sliding windows, statistical parameters (average, standard deviation, \textsc{q\textsubscript{1}}, \textsc{q\textsubscript{2}}, \textsc{q\textsubscript{3}}), signal \textsc{fft} and sensors (see Table \ref{tab:features}). The generation of the sample classification and classification model descriptions are detailed in Algorithm \ref{alg:explainability_sample} and Algorithm \ref{alg:explainability_model}, respectively. Then, the most relevant features and parameters (windows, metrics, and sensors) are ordered by importance and introduced into templates (see Listing \ref{lst:explicability_nl}) to create natural language descriptions automatically. The latter data are also displayed on the user dashboard.

\begin{table*}[tbh]
\centering
\small
\caption{\label{tab:features}Features in the transportation data set.}
\begin{threeparttable}
\begin{tabular}{ccrp{7cm}}
\toprule
\bf Number & \bf Type & \multicolumn{1}{c}{\bf Name} & \multicolumn{1}{c}{\bf Description}\\ \midrule

\multirow{1}{*}{1} & \multirow{10}{*}{Analog} & \multirow{1}{*}{DV ~pressure} & Measures the compressor pressure when the air dryer towers discharge water. \\

2 & & Flowmeter & Measures the airflow circulating in the compressed air circuit.\\

3 & & H1 & Detects when the pressure is above \SI{10.2}{\bar}.\\

\multirow{1}{*}{4} & & \multirow{1}{*}{MC} & Measures the motor current. \\

5 & & Oil temperature & Measures the oil temperature in the compressor.\\

6 & & Reservoirs & Measures the pressure inside the compressed air tanks.\\

7 & & TP2 & Measures the compressor pressure.\\

8 & & TP3 & Measures the pneumatic panel pressure.\\

\cmidrule{2-4}

9 & \multirow{13}{*}{Digital} & Flow rate & Activates when there is a change in the airflow rate.\\

\multirow{1}{*}{10} & & \multirow{1}{*}{COMP} & Activates when there is no air admission in the compressor.\\

\multirow{1}{*}{11} & & \multirow{1}{*}{DV electric} & Activates when the compressor is working under load.\\

12 & & LPS & Activates when the pressure is lower than \SI{7}{\bar}.\\

\multirow{1}{*}{13} & & \multirow{1}{*}{MGP} & Activates when the pressure in the air-producing unit is below \SI{8.2}{\bar}.\\

\multirow{1}{*}{14} & & \multirow{1}{*}{Oil level} & Activates when the oil level of the compressor is below the expected values.\\

15 & & Pressure switch & Activates when there is a discharge in the air drying towers.\\

\multirow{1}{*}{16} & & \multirow{1}{*}{Air dryer tower} & Indicates the tower in operation (zero for tower one, one for tower two).\\
\bottomrule
\end{tabular}
\end{threeparttable}
\end{table*}

\begin{algorithm*}[htb]
 \caption{Explanation of the sample classification.}\label{alg:explainability_sample}
 
\begin{algorithmic}[0]
\small
\Function{features\_involved\_in\_estimator\_decision}{}

 \State node\_data= [\{feature,threshold, left\_branch, right\_branch,type\}] \#List of nodes in the tree path.
 
 \State $sample\_data = [\{feature,feature(value)\}]$ \#Input data sample.

 \State $feature\_relevance=[]$ \#Output feature relevance.
 
 \State $node=node\_data[0]$ \#Starts with the root node.

 \While{$node[type] \neq leaf$}
 
    \State $feature= node[feature]$
 
    \State $threshold = node[threshold]$
	
    \State $right\_branch = node[right\_branch]$
 
    \State $left\_branch = node[left\_branch]$

    \If{$individual\_data[feature(value)] \leq threshold$}
 
	\State $node=left\_branch$
    \Else 
        \State $feature\_relevance.append(feature])$
 
	\State $node=right\_branch$
    \EndIf

 \EndWhile
 
 $feature\_relevance = feature\_relevance.sort\_by\_frequency()$ \#Returns the features involved in the sample classification.
 \EndFunction
\end{algorithmic}
\end{algorithm*}

\begin{algorithm*}[htb]
 \small
 \caption{Explanation of the classification model.}\label{alg:explainability_model}

 \begin{algorithmic}[0]
 
 \Function{Description\_most\_representative\_parameters}{}
 
 $windows = windows.sort\_by\_frequency().get(5) $ 

 $metrics = metrics.sort\_by\_frequency().get(5) $

 $sensors = sensors.sort\_by\_frequency().get(5) $
 
\EndFunction
 \end{algorithmic}
 \end{algorithm*}

\section{Experimental results}
\label{sec:experimental_results}

This section describes the experimental data set (Section \ref{sec:experimental_dataset}) along with the feature engineering (Section \ref{sec:feature_engineering_results}), feature analysis and selection (Section \ref{sec:feature_analysis_selection_results}), classification (Section \ref{sec:classification_results}) and explainability results (Section \ref{sec:explainability_results}). Ultimately, it discusses the results in comparison to those found in the literature (Section \ref{sec:discussion}).

For validation purposes, two experimental scenarios were designed. The first scenario exploits the average and standard deviation of the original features in Table \ref{tab:features}, whereas the second also includes the quartile distribution and \textsc{fft} of the original features obtained with the four \textsc{fir} filters.

\subsection{Experimental data set}
\label{sec:experimental_dataset}

The Metro\textsc{pt} data set \cite{Veloso2022} was gathered at a frequency of \SI{1}{\hertz} from January to June 2022 by the metro operator in Porto, Portugal. The data comprise analog (\textit{e.g.}, current, flow, pressure, and temperature) and digital (\textit{e.g.}, control and status) signals related to the air-producing unit. Table \ref{tab:features} details the features in the Metro\textsc{pt} data set. Each data sample comprises values for the \num{16} features listed. Table \ref{tab:periods} lists the faults within the data set. 

\begin{table}[htb]
\centering
\small
\caption{\label{tab:periods}Failure reports.}
\begin{tabular}{lcc}
\toprule \textbf{Class} & \textbf{Start} & \textbf{End} \\ \midrule
Air leak in the air dryer & 28-02-22 21:53 & 01-03-22 02:00 \\
Air leak in clients & 23-03-22 14:54 & 23-03-22 15:24 \\
Oil leak in the compressor & 30-05-22 12:00 & 02-06-22 06:18 \\
\midrule
\end{tabular}
\end{table}

Only samples from the day before to the day after failures occur are used to ensure rapid evaluation. 
Finally, samples from the two hours preceding the failure were tagged as failures to test whether the proposed approach effectively predicts failures in advance.
The resulting data set holds \num{606424} samples distributed as indicated in Table \ref{tab:dataset_distribution}.

\begin{table}[htb]
\centering
\small
\caption{\label{tab:dataset_distribution}Distribution of classes in the experimental data set.}
\begin{tabular}{lS[table-format=6.0]}
\toprule \textbf{Class} & \multicolumn{1}{c}{\textbf{Number of entries}}\\ \midrule
Non-failure & 372718 \\
Oil leak compressor & 202684 \\
Air leak dryer & 22021 \\
Air leak client & 9001 \\
\midrule
Total & \num{606 424} \\ \bottomrule
\end{tabular}
\end{table}

\subsection{Online data pre-processing}
\label{sec:data_processing_results}

This section details the implementations, design decisions, and results of the feature engineering, feature analysis, and selection stages.

\subsubsection{Feature engineering}
\label{sec:feature_engineering_results}

First, the minimal number of samples (sliding window length) of each \textsc{fir} filter is determined using the first two days of data. The result was \num{1399}, \num{116}, \num{531} and \num{864} samples for the average, \textsc{q\textsubscript{1}}, \textsc{q\textsubscript{2}} and \textsc{q\textsubscript{3}} \textsc{fir} filters, respectively. Next, the system generates, for each data sample, \num{24} new features corresponding to the average, standard deviation, \textsc{q\textsubscript{1}}, \textsc{q\textsubscript{2}}, \textsc{q\textsubscript{3}} and \textsc{fft} features of the four \textsc{fir} filters. The total number of features amounts to \num{400}: \num{16} original features listed in Table \ref{tab:features} plus \num{384} engineered features.

\subsubsection{Feature analysis and selection}
\label{sec:feature_analysis_selection_results}

Feature selection evaluates the cumulative variance of each new feature sample using the \texttt{\small VarianceThreshold} technique (available at \url{https://riverml.xyz/0.11.1/api/feature-selection/VarianceThreshold}, June 2025) from \texttt{\small River} package (available at \url{https://riverml.xyz/0.11.1}, June 2025). The variance threshold was set to \num{0.5}. The selected features per scenario are listed below. For conciseness, the variance values of the selected features are not shown.

\begin{description}

\item \textbf{Scenario 1}: the features engineered from original features 2 to 5, 7, and 8 of Table \ref{tab:features}, corresponding to two new features (average and standard deviation) per selected original feature.
 
\item \textbf{Scenario 2}: the features engineered from original features 1 to 13, and 16 of Table \ref{tab:features}, corresponding to 24 new features (average, standard deviation, \textsc{q\textsubscript{1}}, \textsc{q\textsubscript{2}}, \textsc{q\textsubscript{3}} and \textsc{fft} of the four \textsc{fir} filters) per selected original feature.

\end{description}

In summary, feature selection in both scenarios was carried out using a variance threshold experimentally set at 0.5. This threshold was established after observing signal variations, concluding that a variation below this value can be considered attributable to noise or common sensor behavior, with greater variation being necessary to provide information to the classifier. While signals with lower variations could introduce bias into the prediction, increasing the risk of overfitting in the models, this approach eliminates variables whose variability is insufficient to contribute significantly to classification, reducing dimensionality and promoting model generalization and efficiency. As a result, 12 and 336 features were used in the first and second scenarios, respectively.

\subsection{Online classification}
\label{sec:classification_results}

The stream-based classification explores the following algorithmic implementations: 
\begin{itemize}
 
    \item \textsc{gnb} (available at \url{https://riverml.xyz/dev/api/naive-bayes/GaussianNB}, June 2025).

    \item \textsc{htc} (available at \url{https://riverml.xyz/dev/api/tree/HoeffdingTreeClassifier}, June 2025).
 
    \item \textsc{hatc} (available at \url{https://riverml.xyz/0.13.0/api/tree/HoeffdingAdaptiveTreeClassifier}, June 2025).
 
    \item \textsc{arfc} (available at \url{https://riverml.xyz/dev/api/ensemble/AdaptiveRandomForestClassifier}, June 2025).
 
\end{itemize}

Listings \ref{htc_conf}, \ref{hatc_conf}, and \ref{arfc_conf} present the hyperparameter values explored for the \textsc{htc}, \textsc{hatc}, and \textsc{arfc} models, respectively.  It should be noted that the \textsc{gnb} model does not require hyperparameter configuration since it assumes a Gaussian distribution for each feature and estimates the parameters directly from the observed data. The list of values was selected by searching for a parameter range that would guarantee stable system operation under controlled conditions without increasing server resource consumption. Thus, this exhaustive hyperparameter search process enables each model to be optimally adapted to the monitoring problem, thereby maximizing accuracy while maintaining stability and runtime efficiency. In each case, the values selected as optimal are highlighted in bold.

The applied hyperparameter optimization procedure constitutes an adaptation of the traditional \textit{Grid Search} method (available at \url{https://scikit-learn.org/stable/modules/generated/sklearn.model_selection.GridSearchCV.html}, June 2025), widely used in offline supervised learning. However, since the problem addressed is framed in a streaming classification context, it is not feasible to employ conventional cross-validation since the samples arrive sequentially and cannot be considered independent or interchangeable in time. Therefore, the \textit{Grid Search} scheme was adapted to a \textit{prequential} evaluation framework. This approach enables the system to maintain its online nature, preserving both the temporal integrity of the data and the continuous learning dynamics.

In the case of \textsc{htc}, the tree hyperparameters \textit{depth}, \textit{tie-threshold}, and \textit{max-size} were explored. The best performance was obtained with a maximum depth of 50 nodes, which allows an adequate balance between representation capacity and the risk of overfitting in a continuous learning context, where the number of instances can grow indefinitely. A \textit{tie-threshold} value of 0.5 imposes a relatively conservative margin for decision-making when increases in the gain metric are inconclusive, helping to stabilize tree growth in the advent of minor fluctuations in the data. Finally, a \textit{max-size} of 50 controls the maximum number of memory buffers temporarily stored in each node before performing splits, favoring a sufficiently agile adaptation without resulting in high computational cost.

For \textsc{hatc}, the same ranges of \textit{depth} and \textit{tie-threshold} were used, selecting the optimal values of 50 and 0.5, respectively. However, the \textit{max-size} parameter showed better performance with a value of 100, allowing the model to accumulate more information locally before deciding on structural modifications, which is consistent with the adaptive nature of this algorithm, which seeks to replace underperforming branches dynamically.

For its part, \textsc{arfc} incorporates multiple hyperparameters related to both the ensemble size and the diversification of the base models. The optimal number of trees (\textit{models}) was set to 50, providing a compromise between diversity and computational cost. The \textit{features} parameter also achieved its best performance with 50 features, allowing for an adequate degree of randomization without sacrificing relevant information in each partition. Finally, the \textit{lambda} parameter was set to 50, adjusting the bagging intensity to enhance the internal variability of the ensemble.

\begin{lstlisting}[frame=single,caption={\textsc{htc} hyperparameter configuration (best values in bold).},label={htc_conf},emphstyle=\textbf,escapechar=ä]
depth = [ä\bf50ä, 100, 200]
tiethreshold = [ä\bf0.5ä, 0.05, 0.005]
maxsize = [ä\bf50ä, 100, 200]
\end{lstlisting} 
 
\begin{lstlisting}[frame=single,caption={\textsc{hatc} hyperparameter configuration (best values in bold).},label={hatc_conf},emphstyle=\textbf,escapechar=ä]
depth = [ä\bf50ä, 100, 200]
tiethreshold = [ä\bf0.5ä, 0.05, 0.005]
maxsize = [50, ä\bf100ä, 200]
\end{lstlisting}
 
\begin{lstlisting}[frame=single,caption={\textsc{arfc} hyperparameter configuration (best values in bold).},label={arfc_conf},emphstyle=\textbf,escapechar=ä]
models = [ä\bf50ä, 100, 200]
features = [ä\bf50ä, 100, 200]
lambda = [ä\bf50ä, 100, 200]
\end{lstlisting}

Table \ref{tab:classification_results} provides the classification results obtained in the two experimental scenarios (downsampled by a factor of 50 to reduce over-training with similar samples), with the best results highlighted in boldface. Both scenarios incorporate feature selection and hyperparameter optimization.

\begin{table*}[htbp]
\centering
\small
\caption{\label{tab:classification_results}Online failure detection results (best values in bold, \#1: non-failure, \#2: oil leak, \#3: air leak dryer, \#4: air leak client).}
\begin{tabular}{ccccccccS[table-format=3.2]}
\toprule
\bf \multirow{2}{*}{Scenario} & \bf \multirow{2}{*}{Model} & \bf \multirow{2}{*}{Accuracy} & \multicolumn{5}{c}{\bf \textit{F}-measure} & {\bf Time}\\
\cmidrule(lr){4-8}
 & & & Macro & \#1 & \#2 & \#3 & \#4& {(s)}\\
\midrule
\multirow{4}{*}{1} & 
\textsc{gnb} & 92.19 & 86.16 & 93.41 & 91.67 & 85.01 & \bf 74.55 & 4.95\\
& \textsc{htc} & 77.80 & 64.62 & 83.61 & 70.49 & 53.45 & 50.95 & 6.57\\
& \textsc{hatc} & 94.80 & 77.72 & 95.79 & 96.90 & 73.33 & 44.86 & 5.16\\
& \textsc{arfc} & \bf 97.53 & \bf 88.07 & \bf 98.00 & \bf 98.10 & \bf 95.11 & 61.07 & 135.97\\
\midrule

\multirow{4}{*}{2} &
\textsc{gnb} & 76.15 & 62.62 & 82.54 & 60.90 & 82.33 & 24.69 & 15.36\\
& \textsc{htc} & 71.46 & 73.18 & 79.33 & 47.14 & 89.08 & 77.15 & 15.30\\
& \textsc{hatc} & 95.89 & 90.90 & 96.59 & 96.08 & 88.46 & 82.48 & 21.12\\
& \textsc{arfc} & \bf 99.62 & \bf 98.90 & \bf 99.69 & \bf 99.72 & \bf 98.17 & \bf 98.03 & 209.65\\
\bottomrule
\end{tabular}
\end{table*}

In the first scenario, the results reflect limitations in the classifiers' ability to capture the particularities of minority classes. As a consequence, models such as \textsc{gnb}, \textsc{htc}, and \textsc{hatc} experience a notable drop in performance, as they are not able to characterize more complex aspects of the dynamic behavior of the signals captured by the sensors, such as changes in distribution, the presence of anomalous oscillations or frequency alterations typical of certain types of failures. Even so, the \textsc{arfc} model manages to maintain the best relative performance within this scenario, reaching an accuracy of \SI{97.53}{\percent}, although with room for improvement in class \#4 (air leak client).

In scenario 2, the new features allow the system to capture more complex statistical patterns, anomalies, and frequency components that are not accessible through simple mean and standard deviation measures. Thanks to this richer and more descriptive representation, the models experience a substantial and homogeneous improvement in all the metrics evaluated, with the \textsc{arfc} model reaching its maximum performance with an accuracy of \SI{99.62}{\percent} and \textsc{f}-measure values greater than \SI{98}{\percent} in all classes.

The better performance of \textsc{arfc} is justified by the fact that it is an ensemble of multiple adaptive trees. The \textsc{arfc} has a greater capacity to model nonlinear and complex relationships between features, something especially relevant in a sensor-based system where interference and operation are nonlinear. Furthermore, \textsc{arfc} presents explainability capabilities that are especially valuable in \textsc{p}{\footnotesize d}\textsc{m} applications. Since it is based on decision trees, it is possible to inspect individual decision nodes and the features most frequently involved in classification paths. This traceability is much more limited in other models, such as \textsc{gnb}, where decisions are based on global probability distributions.

Moreover, \textsc{arfc} classifies \num{58} samples per second. Given that the air production unit on board the Metro do Porto trains generates one sample per second, the online classifier performs more than adequately. Furthermore, given the need to sequentially test, train, and evaluate each sample received, the experiments were carried out on a single core of a 20-core platform. In a real deployment, where threads can be assigned to different \textsc{ml} models, the system throughput increases from 58 samples per second (one thread) to 1160 samples per second (20 threads). This scalability potential is essential for modeling complex processes involving multiple subsystems, such as railway networks. Decentralized performance can be addressed by distributing the workload across multiple computing nodes.

\subsection{Online explainability}
\label{sec:explainability_results}

As previously mentioned, the method explains classification outcomes for both \textsc{ml} and maintenance experts. The \textsc{ml} expert explanation details the most relevant features involved in the prediction and the decision paths of each target class (local explanations) along with the overall decision path of the model (global explanations), using natural language (see Listing \ref{lst:explicability_nl}).

\begin{figure*}[htbp]
\centering
\includegraphics[width=0.8\textwidth]{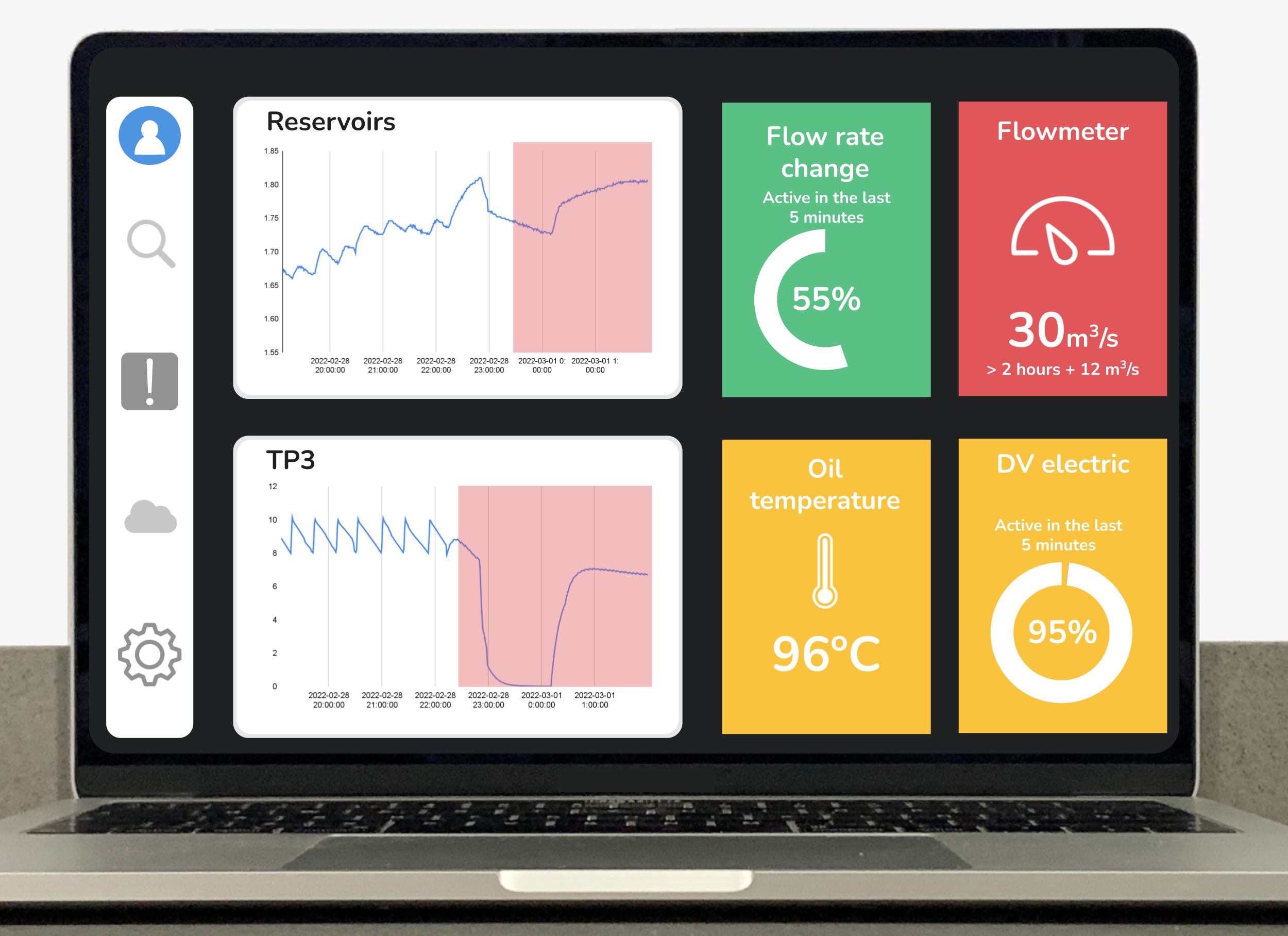}
\caption{\label{fig:dashboard}Explainability dashboard.}
\end{figure*}

The maintenance expert appreciates a clear, natural language explanation and a visual dashboard (see Figure \ref{fig:dashboard}) that highlights the most relevant features in predicting each target class. The dashboard displays the most relevant features' value, status, and normal and abnormal plots (visual and feature-relevance explanations). The \textsc{templated} (available at \url{https://templated.co}, June 2025) library was employed to display the most relevant features and the Highcharts (available at \url{https://www.highcharts.com}, June 2025) for the plots.

\begin{lstlisting}[float=htb,caption={Natural language explanation of a sample classification.},label={lst:explicability_nl}]
For sample 1340, the five most representative features are:
    1. FFT of Reservoirs from Q1-size sliding window
    2. FFT of Reservoirs from Q2-size sliding window.
    3. FFT TP3 from Q3-size sliding window.
    4. FFT of DV pressure from AVG-size sliding window.
    5. FFT of Flowmeter from Q1-sliding window. 

The most representative parameters for the ML model are:
    Sliding windows: four sliding windows contribute equally.
    Signal Pre-Processing Technique: FFT.
    Sensors: Flowmeter, H1, Oil temperature, TP2, TP3.

Given sensors with
    abnormal patterns:
    -  DV pressure sensor (> 30 minutes with significant change)
    -  TP3 (> 15 minutes)
    -  Reservoirs (> 10 minutes)
and anomalous values:
    -  Flowmeter (> 2 minutes)
    -  Reservoirs (> 2 minutes) 
then the prediction is that 
    there is an air leak in the air dryer.
\end{lstlisting}

\subsection{Discussion}
\label{sec:discussion}

It is crucial to elaborate on the practical implications of the results obtained in this study for implementing \textsc{p}{\footnotesize d}\textsc{m} solutions in real-world railway environments. Accordingly, its high performance ensures a robust ability to anticipate critical events, minimizing both false positives and undetected failures. This allows railway operators to significantly reduce unplanned downtime, optimize maintenance planning, and extend component lifespans, resulting in a direct improvement in operational efficiency and a decrease in associated costs. Similarly, the real-time processing of data streams ensures that the solution is compatible with the technical and logistical requirements of the railway environment. In this regard, consideration should be given to the system's ability to classify samples in real time (\textit{i.e.}, 58 samples per second), which demonstrates its feasibility for deployment, even in contexts with multiple subsystems generating data simultaneously. Furthermore, the automatic generation of natural language and visual explanations provides a key differential value from an operational standpoint. This functionality enables maintenance operators to understand the reasons behind a failure prediction, facilitating informed decision-making and reducing diagnosis and resolution times. The latter also increases confidence in the system. Ultimately, the use of a real-world dataset, such as MetroPT, with verified operational failure labels, supports the relevance and applicability of the results. This demonstrates that the proposed solution is not limited to experimental conditions but has been validated in an industrial context with complex, heterogeneous data.

Since the surveyed fault detection/prediction methods use different datasets, comparing the results may not be straightforward. Fumeo, E. \textit{et al.} (2015) \cite{FUMEO2015} employed an \textsc{osvr} to estimate the remaining useful life of components exploring prognostic health monitoring challenge data sets. The solution presents a Mean Absolute Percentage Error (\textsc{mape}) of \SI{2.57}{\percent}. Although decision-making is done online, the authors do not adopt classification techniques. Moreover, the fault detection solution by Ribeiro, R. P. \textit{et al.} (2023) \cite{Ribeiro:2023} processes data online while the classification is performed offline with a Sparse Autoencoder. The evaluation metric used is Root Mean Squared Error (\textsc{rmse}), and the best value obtained is 0.3048. Compared to Fumeo, E. \textit{et al.} (2015) \cite{FUMEO2015} and Ribeiro, R. P. \textit{et al.} (2023) \cite{Ribeiro:2023}, the current solution attains better values (\textit{i.e.}, \SI{0.39}{\percent} \textsc{mape} and 0.0623 \textsc{rmse} for scenario 2 with failure and non-failure categories). In contrast, Meira, J. \textit{et al.} (2023)\cite{Meira2023} proposed a similar solution but without explainability capabilities and lower performance (\textit{e.g.}, \SI{16}{\percent} points lower for \textsc{f}-measure).

The only work that explores the Metro\textsc{pt} data set is that of Davari, N. \textit{et al.} (2021) \cite{Davari2021}. The current solution increases the classification \textit{F}-measure by \SI{194}{\percent}. The performance of the proposed solution and the most closely related work from the literature are summarized in Table \ref{tab:comparisonclassification_results}.

\begin{table}[htbp]
\centering
\small
\caption{\label{tab:comparisonclassification_results}Comparison of online failure detection results using Metro\textsc{pt}.}
\begin{tabular}{ccccccccS[table-format=3.2]}
\toprule
\bf {Proposal} & \bf{\textsc{ml} Technique} & \bf {Accuracy} & \bf \textit{F}-measure & {\bf Time}\\
\midrule
\cite{Davari2021} & Autoencoder & \textsc{na} & 33.60 & \textsc{na}\\
\cite{Meira2023} & XStream & 97.49 & 83.05 & \textsc{na}\\
\midrule
\bf Current & \textsc{arfc} & 99.62 & 98.90 & 209.65\\
\bottomrule
\end{tabular}
\end{table}

Regarding explainability, the current explanations can be compared with those obtained by processing the Metro\textsc{PT} data set with the neural symbolic explainer technique described by Ribeiro, R. P. \textit{et al.} (2023) \cite{Ribeiro:2023}. Listing \ref{rule_expl} holds the explanatory rules of the detected air leak on the train \textsc{apu} where B$i$ indicates the binary representation of analog sensor values and \textsc{re} the autoencoder reconstruction error value. While these explanatory rules constitute a post hoc technique that provides local textual feature-relevance explanations, they are difficult to understand. The current proposal explains fault predictions through global textual feature-relevance explanations and local textual, visual, and feature-relevance descriptions, providing meaningful insights to maintenance experts.

\lstset{keywords={B1_TP3,B2_H1,B5_MC,B6_TP3,Flowrate,RE}}
\begin{lstlisting}[float,caption={Rule-based explanations.},label={rule_expl},emphstyle=\textbf,escapechar=ä]
 Rule 1: ä\textbf{if}ä (B1_TP3 > 7345 ä\textbf{and}ä B5_MC > 1925) ä\textbf{then}ä RE = 1.8116
 Rule 2: ä\textbf{if}ä (Flowrate > 251) ä\textbf{then}ä RE = 2.3932
 Rule 3: ä\textbf{if}ä (B6_TP3 < 5635) ä\textbf{then}ä RE = 2.4445
 Rule 4: ä\textbf{if}ä (B2_H1 > 378) ä\textbf{then}ä RE = 1.8791
\end{lstlisting}

More specifically, our decision path traversing algorithm generates, utilizing natural language models and textual descriptions for the control panel, accompanied by visual elements. Moreover, it analyses the causal pattern of the failure considering the most relevant characteristics (\textit{i.e.}, those whose value exceeds the bifurcation threshold for the failure category) in the decision tree path of the \textsc{arfc} model. In this regard, the oil leak compressor failure is related to high values in the flowmeter (feature 2 in Table \ref{tab:features}), the \textsc{h}1 (feature 3), and the broader wave of the \textsc{mc} (feature 4). In contrast, to detect the air leak dryer failure, high values in the oil temperature (feature 5) and the \textsc{tp}2 (feature 7), and the broader wave of the reservoirs (feature 6) must be considered. Regarding the air leak client failure causal pattern, high oil temperature, and broader wave values for this feature and the \textsc{tp}3 (feature 8) was detected.

Consequently, the explainability of the model enables effective information exchange between humans (\textit{e.g.}, experts, and end-users) and complex, intelligent systems. The railway sector's harsh and dynamic operating conditions make fault prediction a crucial challenge. Specifically, it can significantly reduce the time needed for fault cause analysis by minimizing fault diagnosis tests. The adopted analysis of the most relevant features is well-suited for explaining and troubleshooting predictions. Furthermore, thanks to the generated descriptions, the experts can provide feedback to correct, refine, or even identify additional fault causes. Early failure prediction and explanation improve user experience by reducing service outages and delays.

\section{Conclusion}
\label{sec:conclusion}

This innovative real-time fault prediction provides natural language and visual explainability for \textsc{its}. The ultimate objective of this research is to mine the incoming train sensor data from a public railway operator to anticipate failures and support maintenance decisions.

Thus, the proposed solution detects and explains real-time failures based on the incoming data. The processing pipeline performs: (\textit{i}) online data pre-processing (feature engineering, analysis, and selection), (\textit{ii}) online sample classification, and (\textit{iii}) online sample classification explanation. Consequently, this research advances the field by proposing a transparent and real-time \textsc{p}{\footnotesize d}\textsc{m} solution that incorporates novel features (statistical and frequency-related features) and adopts sliding-window-based processing to create non-stationary online \textsc{ml} models. Moreover, it advances explainable Artificial Intelligence research by presenting comprehensive natural language descriptions and visual indicators in a dashboard for maintenance experts and non-expert users, all in real time.

The experiments were performed with the Metro\textsc{pt} data set, which holds analog and digital data produced by the air-producing unit on board the trains of Metro do Porto. The best results, obtained with the Adaptive Random Forest Classifier, show accuracy as well as macro and micro \textit{F}-measure above \SI{98}{\percent}. Finally, the method explains classification outcomes to help decision-makers understand the underlying problem and trigger maintenance actions. 

Therefore, early detection and explanation of potential failures contribute to enhancing the quality of passenger service and the operator's reputation. Notably, the proposed \textsc{p}d\textsc{m} pipeline accelerates fault detection, troubleshooting, and repair, thereby reducing costs, extending equipment life, and enhancing service quality. The primary weakness of the method lies in the configuration of the process parameters and model hyperparameters. The definition of the window length of the \textsc{fir} filters and the sampling reduction factor used in the feature analysis and selection phase are determined dynamically through iterative adjustment and scoring methods. Due to the resource and time-consuming nature of online hyperparameter optimization, model hyperparameters are set at the beginning of the classification phase. These limitations can be overcome by updating process parameters (periodically) and model hyperparameters (whenever data drifts). Finally, the current fault detection, which works at the individual sensor level, can be more holistic, \textit{e.g.}, taking into account the propagation of anomalies between neighboring sensors. To facilitate future research, additional datasets and language models will be considered to validate the method's applicability and generate more engaging natural language descriptions. Ultimately, in future experiments, we will gather quantitative data related to productivity improvements from the operator to validate our solution's positive impact further.

\bibliography{3_mybibfile}

\section*{Acknowledgements}

This work was partially supported by: (\textit{i}) Xunta de Galicia grants ED481B-2022-093 and ED481D 2024/014, Spain; and (\textit{ii}) Portuguese national funds through FCT -- Fundação para a Ciência e a Tecnologia (Portuguese Foundation for Science and Technology) -- as part of project UIDP/50014/2020 (DOI: 10.54499/UIDP/50014/2020).

\section*{Author contributions}

\textbf{S.G.M.}: Conceptualization, Methodology, Software, Validation, Formal analysis, Investigation, Resources, Data Curation, Writing - Original Draft, Writing - Review \& Editing, Visualization, Funding acquisition. \textbf{F.A.P.}: Conceptualization, Methodology, Software, Validation, Formal analysis, Investigation, Resources, Data Curation, Writing - Original Draft, Writing - Review \& Editing, Visualization, Funding acquisition. \textbf{F.L.}: Conceptualization, Writing - Original Draft, Writing - Review \& Editing. \textbf{B.B.}: Resources, Writing - Review \& Editing. \textbf{B.M.}: Conceptualizations, Writing - Review \& Editing, Supervision, Funding acquisition. \textbf{J.C.B.R}: Conceptualizations, Writing - Review \& Editing.

\section*{Data availability}

The data is publicly available in the work by Veloso, B. \textit{et al.} (2022) \cite{Veloso2022}.

\section*{Competing interest}

The authors declare no competing interests.

\end{document}